%% file: ms.tex
\documentclass[preprint,12pt]{elsarticle}

\usepackage{amssymb}
\usepackage{amsmath}

\usepackage{mathtools}

\usepackage{algorithm}
\usepackage{algpseudocode}

\usepackage{tabularx}

\usepackage{verbatim}

\usepackage[printonlyused]{acronym}
\usepackage[hang,flushmargin]{footmisc}

\usepackage{hyperref}

\newcommand\blfootnote[1]{%
  \begingroup
  \renewcommand\thefootnote{}\footnote{#1}%
  \addtocounter{footnote}{-1}%
  \endgroup
}
\usepackage{subfig}

\biboptions{numbers,sort&compress}

\begin{document}

\begin{frontmatter}
\title{CARE to Compare: A real-world dataset for anomaly detection in wind turbine data}

\author[address1]{Christian G{\"u}ck\corref{cor1}}
\ead{christian.gueck@iee.fraunhofer.de}
\author[address1]{Cyriana M.A. Roelofs}
\ead{cyriana.roelofs@iee.fraunhofer.de}
\author[address1]{Stefan Faulstich}
\ead{stefan.faulstich@iee.fraunhofer.de}
\cortext[cor1]{Corresponding author}

\address[address1]{Fraunhofer IEE, Joseph-Beuys-Straße 8, 34117 Kassel, Germany}

\begin{abstract}
\input{abstract}
\end{abstract}


\end{frontmatter}


\section{Introduction}\label{introduction}
\input{introduction}

\blfootnote{Nomenclature:
\begin{acronym}
\acro{wt}[WT]{wind turbine}
\acro{ad}[AD]{anomaly detection}
\acro{ml}[ML]{machine learning}
\acro{mse}[MSE]{mean square error}
\acro{nbm}[NBM]{normal behaviour model}
\acro{scada}[SCADA]{Supervisory Control and Data Acquisition}
\acro{ws}[WS]{weighted score}
\acro{acc}[Acc]{accuracy-score}
\acro{om}[O\&M]{Operation \& Maintenance}
\acro{care}[CARE]{Coverage Accuracy Reliability Earliness}
\acro{nn}[NN]{neural network}
\acro{ae}[AE]{autoencoder}
\acro{re}[RE]{reconstruction error}
\acro{auc}[AUC]{area under the curve}
\acro{roc}[ROC]{receiver operating characteristic curve}
\end{acronym}
}

\section{Related work}\label{section related_work}
\input{related_work}

\section{Data}\label{section data}
\input{data_intro}
\subsection{Data requirements} \label{section requirements data}
\input{requirements}
\subsection{Dataset}\label{section dataset}
\input{data}
\subsection{Data labeling}\label{section labeling}
\input{labeling}
\subsection{Anonymization} \label{section anonymization}
\input{anonymization}

\section{Anomaly detection evaluation}\label{section evaluation}
\input{evaluation}
\subsection{The CARE-Score} \label{section score}
\input{score}
\subsection{Mini-Benchmark}\label{section benchmark}
\input{benchmark}

\section{Summary}\label{section summary}
\input{summary}

\paragraph{Acknowledgements}
The development of methods presented was funded by the German Federal Ministry for Economic Affairs and Climate Action (BMWK).

\bibliographystyle{elsarticle-num}
\bibliography{references}

\end{document}

%% file: abstract.tex
Anomaly detection plays a crucial role in the field of predictive maintenance for wind turbines, yet the comparison of different algorithms poses a difficult task because domain specific public datasets are scarce.
Many comparisons of different approaches either use benchmarks composed of data from many different domains, inaccessible data or one of the few publicly available datasets which lack detailed information about the faults. Moreover, many publications highlight a couple of case studies where fault detection was successful.
With this paper we publish a high quality dataset that contains data from 36 wind turbines across 3 different wind farms as well as the most detailed fault information of any public wind turbine dataset as far as we know. The new dataset contains 89 years worth of real-world operating data of wind turbines, distributed across 44 labeled time frames for anomalies that led up to faults, as well as 51 time series representing normal behavior. Additionally, the quality of training data is ensured by turbine-status-based labels for each data point.
Furthermore, we propose a new scoring method, called CARE (Coverage, Accuracy, Reliability and Earliness), which takes advantage of the information depth that is present in the dataset to identify a good all-around anomaly detection model. This score considers the anomaly detection performance, the ability to recognize normal behavior properly and the capability to raise as few false alarms as possible while simultaneously detecting anomalies early.

%% file: introduction.tex
Wind energy plays a crucial role in the transition to renewable energy, but monitoring and maintaining wind farms and turbines is a costly challenge. These farms are often located in regions with challenging weather conditions, leading to complex operating conditions and increased risk of unexpected failures and downtime. Over the past decade, various approaches for condition monitoring, many of which focus on early fault detection using \ac{scada} data, have been investigated \cite{tautzweinert_using_2017,helbing_deep_2018,pandit2024}.

A common method to detect component failures early is \ac{ad}, which identifies outliers or other anomalous patterns in the data. In the context of \acp{wt}, most \ac{ad} techniques utilize data from the \ac{scada} system, failure logs, vibration data and occasionally status and maintenance logs \cite{latiffianti_wind_2022}. This paper specifically focuses on \ac{ad} models based on \ac{scada} data which are validated using additional failure information.

While there have been several benchmarks \cite{adbench2022, numenta2015}, reviews \cite{pang2022} and comparisons \cite{schmidl_anomaly_2022} of general \ac{ad}-algorithms, most of them use data from a wide variety of domains like spacecraft, medical applications and IT-related data. However, efforts on wind energy specific \ac{ad} are usually based on non-public inaccessible data. For example \cite{zhang_research_2024, yang_luoxiao_conditional_2021} use inaccessible data from wind farms that are located in China, and \cite{morrison_anomaly_2022, schroder_using_2022, mckinnon_comparison_2020} use data from anonymized offshore wind farms. These are 5 recently published examples, which lack the ability for meaningful comparisons between the proposed fault detection algorithms. Also, inaccessible data prevents reproducibility of presented results. 
Some studies have used public wind energy datasets (for example \cite{jia_condition_2021, latiffianti_wind_2022, de_sa_wind_2020, udo_data-driven_2021, tang_fault_2023, jankauskas_exploring_2023, barber_best_2023, barber_enabling_2022}), but they lack comprehensive information about anomalies or component faults. 
The lack of extensive public datasets with both \ac{scada} time series and failure information is a significant limitation in the field of \ac{wt} \ac{scada} data analysis. To enable meaningful comparisons between AD algorithms in the wind energy domain, new public benchmark datasets are necessary.

The main contribution of our work is the publication of the most extensive \ac{wt} \ac{scada} dataset\footnote{The data can be found on ``Fordatis'': \url{http://dx.doi.org/10.24406/fordatis/343} and on ``Zenodo": \url{https://zenodo.org/doi/10.5281/zenodo.10958774}} for \ac{ad} yet. This includes high dimensional data from multiple wind farms, information about the \ac{wt} status at all times, labeled anomalies with annotated starts and ends and additional fault descriptions. Because the data stems from real-world operating wind farms it had to be anonymized with the focus on minimizing the loss of useful information and maximizing the meaningfulness of this dataset for \ac{ad} and predictive maintenance.

In addition to the dataset we also provide a sophisticated score, the \ac{care}-score, for evaluating \ac{ad}-algorithms on this and similar datasets. This score takes into account four key aspects of a high-quality \ac{ad} model for predictive maintenance. In combination with the dataset this score provides the possibility to compare a variety of different \ac{ad}-algorithms, from unsupervised to semi-supervised techniques, designed for early fault detection in \ac{wt}.

The content of this paper divides into the following sections. At first we give an overview about the related work in section \ref{section related_work}. After that we introduce the dataset in section \ref{section data} by giving information about the layout, the requirements we set for the quality of the data, the labeling process and the anonymization actions that were taken. Following this we provide our scoring idea together with a mini-benchmark of a few selected \ac{ad}-algorithms in section \ref{section evaluation}. Finally a summary concludes this work in section \ref{section summary}.

%% file: related_work.tex
In the field of \ac{ad} benchmark data, many studies focus on dataset compositions from various different domains and use cases. Many benchmark datasets also include a mix of artificial data and data from real-world applications. While \cite{adbench2022, pang2022, nassif_machine_2021, ruff_unifying_2021} study a wide spectrum of different \ac{ad} algorithms for a broad collection of data types, there are also several \ac{ad} benchmarks which focus on time series data. One example of such benchmarks is the widely used and cited Numenta benchmark \cite{numenta2015}, which provides a collection of datasets. A more recent and more comprehensive evaluation of \ac{ad} in time series is found in \cite{schmidl_anomaly_2022}, where over 71 algorithms were evaluated on more than 900 time series. 

In the scope of \ac{ad} for \acp{wt} the time series datasets mentioned above are usually too broad to be used for evaluation in this specific context. Also, many are either univariate or synthetic time series and therefore not applicable to \ac{ad} in \ac{scada}-data. Unfortunately, most domain-specific evaluations are conducted on inaccessible data that were provided only for the research in which they are used. As mentioned in the introduction, there are plenty examples of studies which use such datasets. 

There are only a handful of open datasets containing \ac{wt} \ac{scada}-data. The studies \cite{effenberger_collection_2022, menezes_wind_2020} give an overview about existing datasets for \acp{wt}. Additionally the Git-repository \cite{letzgus_wind_nodate} summarizes some currently existing datasets although some of the listed datasets, such as the \ac{scada}-data of the ENGIE wind farm ``La Haute Borne", are not available anymore. The most relevant public dataset in the context of \ac{ad} for early fault detection is provided by the EDP open data platform \cite{edp_inovacao_edpr_2018}, since it is, as far as the authors know, the only one containing information about \ac{wt} faults in addition to the \ac{scada}-data. The faults are provided in form of a start timestamp for some turbine faults.
This dataset was used in the fault detection challenge ``hack the wind" \cite{hack_the_wind_2018} hosted by EDP which is mentioned and evaluated together with the ``WeDoWind"-challenge \cite{we_do_wind_challenge_2021} in \cite{barber_enabling_2022} and \cite{barber_best_2023}. These challenges focus in particular on the evaluation of \ac{ad}-algorithms based on maintenance cost and potential savings that could be achieved through predictive maintenance. Furthermore, several studies on fault detection have used this data \cite{latiffianti_wind_2022, de_sa_wind_2020, tang_fault_2023, jankauskas_exploring_2023}. Although the EDP-dataset is widely used, its level of detail regarding the fault information is small, especially in comparison to the inaccessible datasets mentioned before.

The lack of publicly available \ac{scada}-datasets of \acp{wt} is also acknowledged in \cite{pandit2024}, which highlights that it is a constraint in the progress of \ac{wt} \ac{scada} applications. Additionally, the absence of publicly available datasets containing real-world anomalies is recognized as a significant obstacle in the development of \ac{ad} in general, as it may not adequately reflect the performance of methods in real-world applications \cite{pang2022, ruff_unifying_2021}. 

But there is not only need for additional domain specific public datasets, the data quality and the level of detail also plays an important role. As pointed out by \cite{renjie2023} many \ac{ad} benchmark datasets suffer from flaws that limit their significance. The main flaws are defined as the flaw of ``Triviality", ``Unrealistic Anomaly Density", ``Mislabeled Ground Truth" and the ``Run-to-Failure Bias". In the case of publicly available wind \ac{scada} datasets, one common issue is the absence of labels, particularly regarding fault information.
\newline

Considering the flaws in datasets, scoring for \ac{ad} algorithms poses a difficult challenge. Many studies utilize standard classification metrics such as accuracy, precision, recall, or the \ac{auc} of the \ac{roc} \cite{adbench2022, morrison_anomaly_2022, chen_anomaly_2021, mckinnon_comparison_2020}.

While it is possible to evaluate \ac{ad} algorithms using the \ac{auc}-\ac{roc} score for all possible thresholds, for most practical applications it is much more useful to have a high F-Score, or a related score. In \cite{astha2022} several variants of F-scores are compared. The standard pointwise F-Score is the simplest, but for most use cases, the interest lies in detecting anomaly events, i.e., a continuous set of anomalous time points, rather than individual time points. A composite F-score is introduced, a modification of the classic F-score that takes into account anomaly events through event-wise recall.

Another approach, presented in \cite{carrasco2021}, modifies the classic \ac{auc}-\ac{roc} metric by generalizing the concept of the \ac{roc} to the Preceding-Window-\ac{roc}, thereby adjusting the measure to better fit \ac{ad} evaluations on time series data from an event-based perspective.

Finally, the Numenta Benchmark \cite{numenta2015} defines a score that is supposed to measure the performance of more general \ac{ad} models for time series data across different domains. The score is based on 5 key-aspects of a good \ac{ad} model: ``detection of all anomalies", ``early detection of anomalies", ``no false alarms", ``uses only real time data" and ``automation across all different datasets".
\newline

Based on the provided overview of related work, this paper contributes to the progress of \ac{ad} for predictive maintenance on \acp{wt} by introducing a new public dataset that offers more detailed information about turbine faults and associated anomalies. Furthermore, the new dataset addresses the flaws identified by \cite{renjie2023}, although the potential for mislabeled ground truth cannot be completely eliminated in this context, as the start of anomalous behavior is often unclear.
The flaw of triviality is tackled by the inclusion of complex anomalies from real-world \acp{wt} based on feedback of the wind farm operators. Additionally, the proposed \ac{care}-score, which differs from standard classification metrics, draws inspiration from the first three key aspects of the Numenta score and the composite F-Score from \cite{astha2022}, while distinct adaptions and further developments have been made to better fit the specific use case of \ac{ad} for predictive maintenance on \acp{wt}.

%% file: data_intro.tex
In this section, we describe the new dataset provided with this paper. First, we discuss the requirements for a good dataset for \ac{ad} in \acp{wt} in section \ref{section requirements data}. Then, we provide an overview of the data published in section \ref{section dataset}, including general statistics such as the number of anomalous events and features, as well as data quality. In section \ref{section labeling}, we explain the process of labeling each time series and datapoint, and in section \ref{section anonymization} the anonymization process is described.

%% file: requirements.tex
During the process of selecting data for this benchmark dataset, seven requirements were defined to ensure the quality and significance of comparisons of \ac{ad} algorithms for \ac{ad} in \acp{wt}. The requirements are as follows: 

\begin{enumerate} \label{enumerate: data requirements}
    \item The dataset must contain as many anomaly events as possible.
    \item The dataset must contain different wind farms.
    \item The dataset must contain different fault types.
    \item The dataset must be balanced, i.e. contain enough prediction data representing normal behavior.
    \item Every sub-dataset must contain enough normal behavior data in the intended training time frame. If at least 2/3 of the training data are normal behavior data we define the sub-dataset to be sufficient.
    \item Every sub-dataset must contain at least one whole year worth of data, to be able to learn seasonality-related effects.
    \item Every anomaly must have an assigned start timestamp. The anomaly end is the start of a turbine fault.
\end{enumerate}

While requirements 1 to 3 are necessary to test the generalization ability of \ac{ad} algorithms, requirement 4 enables tests for the ability to learn normal behavior effectively. This is particularly important for the evaluation of \acp{nbm}. Additionally, requirements 5 and 6 ensures the quality of the training data, to guarantee an \ac{nbm} can be trained. Finally, requirement 7 allows for the evaluation of \ac{ad} models using classification measures. These requirements ensure that the dataset is of high-quality, comprehensive and balanced to train a proper \ac{nbm}, with detailed labels to validate the model. All these properties are also relevant for the definition of the score introduced in section \ref{section score}.

%% file: data.tex
The data consists of 95 datasets, containing 89 years of \ac{scada} time series distributed across 36 different \acp{wt} from the three wind farms A, B and C. The data for Wind farm A is based on the earlier mentioned EDP-data \cite{edp_inovacao_edpr_2018}, and consists of 5 \acp{wt} of an onshore wind farm in Portugal. From this data 22 datasets were selected to be included in this data collection. The other two wind farms are offshore wind farms located in Germany. All three datasets were anonymized as described in section \ref{section anonymization}.
The overall dataset is balanced, as 44 out the 95 datasets contain a labeled anomaly event and the other 51 datasets represent normal behavior.
Each dataset is provided in form of a csv-file with columns defining the features and rows representing the data points of the time series.

The datasets consist of \ac{scada} time series data for each turbine, with a resolution of 10 minutes. Each dataset includes one year worth of data for training a model, as well as 4 to 98 days of prediction data.

The prediction data is divided into an event time frame, with varying amounts of padding data before and after the event. This padding is used to prevent guessing the event label (``anomaly'' or ``normal'') based on the amount of prediction data.

The number of features in the datasets varies depending on the wind farm. Wind farm A has 86 features, wind farm B has 257 features, and wind farm C has 957 features.
In addition to the sensor data features, each time series includes 5 descriptive features: a row ID and a timestamp, an asset ID that identifies the \ac{wt}, a ``train\_test" column indicating whether the row belongs to the training or prediction data, and a status-ID indicating the turbine status at the timestamp.

The remaining features represent sensor measurements. For each sensor, the 10-minute average value is available. Some sensors also have additional information in the form of 10-minute minimum, maximum, and standard deviation values.
The original sensor names have been replaced in order to anonymize the data, as described in section \ref{section anonymization}. Only features that describe power, reactive power or wind speed are recognizable by their name. To accommodate for the loss in information, additional descriptions are provided for every sensor. These descriptions include a brief text, the unit of the sensor as well as boolean indicators that imply whether the sensor represents a regular sensor signal, a counter or an angle.
The most important statistics of the data are summarized in table \ref{table data stats}. The rows ``Anomaly events" and ``Normal behavior" describe the number of datasets containing an anomaly event and without anomalies respectively.

Regarding the data quality there are two challenges. The data for wind farm B and C was provided by the operator with 0-values replacing all missing values, so large amounts of consecutive 0-values must be treated with caution.
Secondly, note that the status values for wind farm B and C may be inconsistent; often the status is only logged when it changes, which may fail if there is a brief communication error. Also, the status values for wind farm A were derived based on the EDP fault logbook, which only contained start timestamps of the faults (see section \ref{section labeling}). It is therefore advisable to check the power and wind speed values in addition to the status values to determine whether the turbine has indeed been operating normally.

\begin{center}
\begin{tabularx}{\textwidth}{ 
    >{\raggedright\arraybackslash}X 
  | >{\centering\arraybackslash}X
  | >{\centering\arraybackslash}X 
  | >{\centering\arraybackslash}X 
  | >{\centering\arraybackslash}X 
} \label{table data stats}
    
    & Wind Farm A & Wind Farm B & Wind Farm C & Overall \\ 
    \hline
    Turbines & 5 & 9 & 22 & 36\\ 
    \hline
    Datasets & 22 & 15 & 58 & 95\\
    \hline
    Anomaly events & 11 & 6 & 27 & 44\\
    \hline
    Normal behavior & 11 & 9 & 31 & 51\\
    \hline
    Features & 86 & 257 & 957 & -\\
    \hline
    Sensors & 54 & 63 & 238 & -\\
\end{tabularx}
\end{center}

%% file: labeling.tex
The data is labeled on two levels. The first level are the so-called event labels. If a dataset contains an anomaly event inside the prediction time frame, the dataset is labeled as an anomaly. If this is not the case it is labeled as normal. The anomaly labels have been determined either based on direct feedback by the wind farm operators or based on documented faults in the form of service reports and fault logbooks.
The normal labels have been determined by a combination of feedback of the wind farm operators, manual inspection of the data and expert knowledge.

For wind farm A all anomaly event starts were defined based on the available EDP fault logbook which only defines start timestamps for each fault. Since no further information is available, analysis of the data before every fault was used to determine possible event starts. The 'true' anomaly event starts for wind farm A can differ from the set ones.

For the wind farms B and C all starts of the anomaly events were defined based on data analysis, feedback of the wind farm operator, service report documents and expert knowledge. While the true starts of the anomaly events could potentially differ from the set ones in some cases, it is highly unlikely that the defined events start too early. If anything, anomaly event start could be earlier than defined.

The second level of labeling assigns a label to each timestamp of every dataset. These labels are called status-IDs. For the wind farms B and C they are derived from the original operating modes that were provided by the wind farm operators in combination with service report information. For wind farm A this information was not provided. 
In this case the status-IDs were based on the fault information from the logbook provided by EDP. For each turbine fault the preceding 14 days were marked with the status-ID 4 (fault) and the 3 days after the fault timestamp were marked with the status-ID 3 (service mode). The time ranges around the turbine fault were set with the aim in mind to reduce the risk of including anomalous behavior in the training data. As no information is available on the duration of anomalies before and after the given faults, the time ranges were chosen conservatively. 

The status labels can be used to infer whether a given data point represents normal \ac{wt} behavior or not. The status-IDs, their description and whether we consider the status normal, are found in table \ref{table status}.

\begin{center}
\begin{tabularx}{\textwidth}{
    >{\centering\arraybackslash}X
  | >{\centering\arraybackslash}X 
  | >{\centering\arraybackslash}X
} \label{table status}
    Status-ID & Description & Considered Normal  \\
    \hline
    0 &	Normal operation without limitations & True \\
    \hline
    1 & Derated power generation with a power restriction & False \\
    \hline
    2 &	Asset is idling and waits to operate again	& True \\
    \hline
    3 & Asset is in service mode / service team is at the site & False \\
    \hline
    4 & Asset is down due a fault or other reasons & False \\
    \hline
    5 & Other operational states for example system test, setup, ice build-up or emergeny power & False
\end{tabularx}
\end{center}

%% file: anonymization.tex
Due to confidentiality reasons the data of wind farm B and C was anonymized. The anoymization includes the removal of all information that can directly identify the wind farms, such as the name of the wind farm, the original names of each \ac{wt}, the turbine type and the location. The wind farm names were replaced by the generic names "Wind Farm A", "Wind Farm B" and "Wind Farm C" while the \ac{wt} names were replaced by randomized asset IDs. However, the asset-IDs were assigned in a way that makes it still possible to link different datasets that belong to the same \ac{wt}.

Additionally, the timestamps of each dataset were shifted by a random number of years. This preserves the consistency of the seasonal information, although it does distort the temporal order of the datasets.

Names of the original SCADA-features were replaced by a numeration of the features. Only features that describe power, reactive power or wind speed are recognizable by their name. Additionally, power and reactive power features have been scaled with the rated power of the turbine. This way, it is still possible to clean and analyse the data using the power curve of the \ac{wt}.

All status information were aggregated from the original status data of the \acp{wt} and the name of each status condition was replaced by a number in combination with a brief description. Wind farms B and C contain detailed status information while wind farm A only contains status information which indicate turbine faults. 

%% file: evaluation.tex
Evaluation of \ac{ad} algorithms poses a difficult task. On one hand the perfect \ac{ad} should detect all anomalies as soon as possible, without any false alarms, on the other hand labeling of anomalies and finding proper start and end times of anomaly events cannot be done perfectly. 

Although the ground truth of every \ac{ad} evaluation is almost certainly flawed \cite{renjie2023}, standard classification metrics, like the F-Score, accuracy and precision, are often used to measure the performance of \ac{ad} algorithms to compare them with other algorithms or to show their overall performance \cite{stetco_machine_2019}.
The F-score in particular is widely used, but it cannot be applied to evaluate the performance of \ac{ad} algorithms on normal data since true negatives are not considered in the F-score.
This is one of the reasons why metrics like the F-score are not suitable for a complete evaluation.

To tackle these problems, we introduce the \ac{care}-score in \ref{section score} to evaluate models on the dataset described in \ref{section data}. The score is composed out of four sub-scores, each evaluating a key aspect of a good \ac{ad} model. In addition to that, we conduct a mini-benchmark \ref{section benchmark} to showcase the \ac{care}-score and dataset.

%% file: score.tex
In the context of \ac{ad} for predictive maintenance the performance of models is often difficult to assess. To address this, we introduce the \ac{care}-score for evaluating \ac{ad} in an operational predictive maintenance setting. The \ac{care}-score focuses on four key aspects that a good AD model for predictive maintenance should excel in, which are:

\begin{enumerate} \label{score properties}
    \item Coverage: Detection of as many correct anomalies as possible,
    \item Accuracy: Recognition of normal behavior,
    \item Reliability: Few false alarm events,
    \item Earliness: Detection of anomalies before fault gets critical.
\end{enumerate}

The \ac{care}-score consists of four sub-scores, each representing one of the four aspects mentioned above. 
The first and fourth sub-scores measure the pointwise classification performance of an algorithm on datasets where anomaly events are present. The second sub-score considers the model's performance on datasets without any anomalous data, i.e. its ability to recognize normal behavior accordingly. The third sub-score assesses classification performance on aggregated events, by applying an eventwise classification measure.

Contrary to measures like the \ac{auc}-\ac{roc} all four sub-scores considered are threshold-specific performance measures. This is a property that makes comparisons of algorithms for the sake of \ac{ad} less significant, but in exchange it raises the significance of the score in a real-world operative predictive maintenance setting. As mentioned in \cite{astha2022}, operators of wind farms and other assets are primarily interested in accurately detecting anomalies while minimizing false alarms. Additionally, this enables comparison of the same \ac{nbm} with different threshold methods.

\subsubsection{Score Definition}

\paragraph{\textbf{Coverage}} \label{paragraph coverage}
In order to measure the coverage - i.e. the classification performance on a time series with anomalies - the $F_\beta$-score is used. At first the prediction time frame of a given anomaly event dataset is filtered. All data points with an abnormal status-ID according to table \ref{table status} are ignored. This is important because these data points are usually very easy to detect. Moreover, the wind farm operator is already informed about the abnormal behavior through the status-ID. Thus these data points are irrelevant in the context of predictive maintenance and they would dilute the score. Now let $\mathbf{g}$ be the ground truth of all data points with a normal status-ID within the prediction time frame and $\mathbf{p}$ is the corresponding prediction of an \ac{ad}-model. The $F_\beta$-score is then defined as
\begin{align} \label{equation f-score}
    F_\beta(\mathbf{g}, \mathbf{p}) = \frac{(1+\beta^2) \cdot tp(\mathbf{g}, \mathbf{p})}{(1+\beta^2 \cdot tp(\mathbf{g}, \mathbf{p}) + \beta^2\cdot fn(\mathbf{g}, \mathbf{p}) + fp(\mathbf{g}, \mathbf{p}))},
\end{align}
where $tp(\mathbf{g}, \mathbf{p})$ is the number of true positives based on $\mathbf{g}$ and $\mathbf{p}$, $fn(\mathbf{g}, \mathbf{p})$ is the number of false negatives and $fp(\mathbf{g}, \mathbf{p})$ is the number of false positives.
In this case, a value of $\beta = \frac{1}{2}$ is chosen to give more weight to precision than recall, thereby penalizing excessive false positives.
\newline

\paragraph{\textbf{Accuracy}}
To measure the performance on datasets that exclusively contain normal behavior, the \ac{acc} is used. Since there are no true positives for the prediction time frames of those datasets, $F_\beta$ from equation \ref{equation f-score} would always be 0. With the same reasoning as in the coverage-paragraph \ref{paragraph coverage} only data points with a normal status-ID are relevant. Let $\mathbf{g}$ be the ground truth of all data points with a normal status-ID within the prediction time frame and $\mathbf{p}$ is the corresponding prediction. Then \ac{acc} is calculated by
\begin{align}
    Acc(\mathbf{g}, \mathbf{p}) = \frac{tn(\mathbf{g}, \mathbf{p})}{fp(\mathbf{g}, \mathbf{p}) + tn(\mathbf{g}, \mathbf{p})}.
\end{align}
where $tn(\mathbf{g}, \mathbf{p})$ is the number of true negatives based on $\mathbf{g}$ as well as $\mathbf{p}$ and $fp(\mathbf{g}, \mathbf{p})$ is the number of false positives.  Note that in contrast to the standard accuracy, there are no true positives and false negatives, since only datasets containing normal behavior are considered and data points with an abnormal status-ID are excluded.
\newline

\paragraph{\textbf{Reliability}}
False alarms on an event basis are taken into account by the event based $F_\beta$-score ($EF_\beta$).
First, each time series prediction needs to be classified as either `anomaly event detected' or `normal behavior'. For this, we first calculate the maximum `criticality', which is a counter-like measure. 
Given the prediction timestamps $t_1,\dots,t_N$, the status information $s_{t_i} \in \{0,1\}$ where $1$ corresponds to a normal status and $0$ to an abnormal status and the prediction $p_{t_i} \in \{0,1\}$ where $1$ represents a detected anomaly and $0$ no detected anomaly for $i=1,\dots,N$, the criticality is computed by algorithm \ref{algorithm criticality}.
\begin{algorithm}
\caption{Criticality Algorithm}\label{algorithm criticality}
\begin{algorithmic}
\State $crit \gets [0,0,\dots,0]\in\mathbb{N}^{N+1}$
\For{$i\in\{1,\dots, N\}$}
\If{$s_{t_i} = 0$} 
    \If{$p_{t_i} = 1$} 
        \State $crit[i] \gets crit[i-1] + 1$
    \Else
        \State $crit[i] \gets \max\{crit[i-1] - 1, 0\}$
    \EndIf
\Else
    \State $crit[i] \gets crit[i-1]$
\EndIf 
\EndFor
\State $crit \gets crit[1:N]$
\end{algorithmic}
\end{algorithm}

After calculating the criticality for the entire prediction time frame, the maximum criticality is compared to a threshold $t_c$, which we set to 72. The idea behind this threshold is that in order to reach a criticality of 72 the algorithm must either detect at least 72 anomalies in a row, which equates to 12 hours of consecutive anomalies, or even more anomalies in the case of non-consecutive detected anomalies. Setting the threshold at 72 was found to be  the most appropriate for all 95 time series in this dataset and generally depends on the length of the time series and the use-case specific definitions of detected anomaly events. For shorter events, a lower threshold is more appropriate.

If the threshold is exceeded, the prediction is counted as a detected anomaly event (i.e. an alarm was raised). If the maximum criticality is below 72, the prediction is counted as a normal event prediction (i.e. no alarm).
These event prediction labels are then compared to the true dataset labels and the $F_\beta$-score is calculated as defined in equation \ref{equation f-score} where $\beta=\frac{1}{2}$ is usually chosen to penalize false positives further.  
\newline

\begin{figure}
    \centering
    \includegraphics[width=\textwidth]{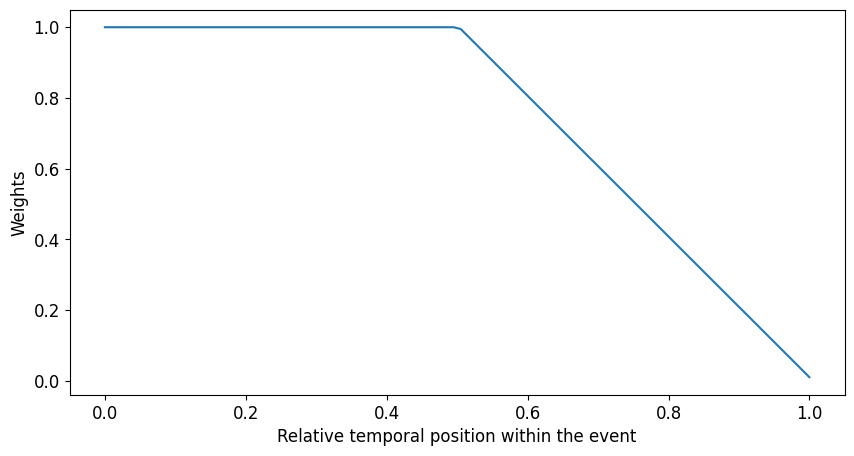}
    \caption{Weight function of the weighted score (WS) for early anomaly detection.}
    \label{figure weight function}
\end{figure}

\paragraph{\textbf{Earliness}}
Similar to the $F_\beta$-score, the second sub-score - the \ac{ws} - is also only applied to anomaly events. As a modified version of the weighted score defined in the Numenta benchmark \cite{numenta2015}, this score weights detected anomalies during the beginning of defined anomaly events higher than detected anomalies at the end of the event time frame. However, instead of discarding additional detected anomalies within the event time frame, all detected anomalies will be considered with positive weights. The piecewise linear function shown in figure \ref{figure weight function} is used to weight the predicted timestamps. In the first half of the event, all detected anomalies are assigned a weight of 1. In the second half of the event, the weights decrease linearly to 0, as the detected anomalies become less important to the wind farm operator the closer they are to the actual turbine fault.

In order to apply this weighting function to anomaly events of different lengths, the length of each event is used to convert the event timestamps to the relative position in the interval $[0,1]$, where $0$ corresponds to the beginning of the anomaly event and $1$ corresponds to the end of the event, i.e. the start of a downtime or a fault as detected by other systems.
Let the consecutive timestamps of an anomaly event $a$ be $t_1<t_2<\dots<t_M$ while $\mathbf{p}_a\coloneqq(p_{t_1},\dots, p_{t_N})\in \{0,1\}^N$ denotes the corresponding prediction where 1 marks a detected anomaly and 0 means no detected anomaly. The \ac{ws} of this anomaly event is then calculated as
\begin{align}
	WS(\mathbf{p}_a) = \frac{\sum_{i=1}^M w_{t_i} \cdot p_{t_i}}{\sum_{i=1}^M w_{t_i}}
\end{align}
where $w_{t_i}\in[0,1]$ is the weight for the timestamp $t_i$.

\paragraph{\textbf{CARE}}
Finally, the \ac{care} score is calculated by combining the four sub-scores. This is done by calculating the event based score $EF_\beta$ and the averages $\overline{F_\beta}$, $\overline{WS}$ and $\overline{Acc}$. Here, $\overline{F_\beta}$ is the arithmetic mean over all $F_\beta$-scores of datasets containing an anomaly event, $\overline{WS}$ is the arithmetic mean over all \ac{ws} of datasets containing an anomaly event and $\overline{Acc}$ is the average over all \ac{acc} of datasets representing normal behavior.

The final \ac{care} score takes two special cases into account. If no anomalies were detected at all, the \ac{care} score will be 0. Also, if $\overline{Acc}$ falls below $0.5$, the predictions are worse than uniformly distributed random predictions. In this case the final score will be equal to $\overline{Acc}$. Outside of these two special cases, the \ac{care} score is defined by a weighted average $WA$ of all sub-scores:
\begin{align} \label{equation: weighted average}
    WA \coloneqq \frac{1}{\sum_{i=1}^4\omega_i}\left(\omega_1 \overline{F_\beta} + \omega_2 \overline{WS} + \omega_3 EF_\beta + \omega_4 \overline{Acc}\right),
\end{align}
where we choose $\omega_1=\omega_2=\omega_3=1$ and $\omega_4=2$ in order to weight the normal datasets to the same degree as datasets containing an anomaly event and $\beta=\frac{1}{2}$.

To summarize, the \ac{care}-score is defined by
\begin{align} \label{equation: care}
CARE \coloneqq 
\begin{cases}
    0, & \text{if no anomalies were detected}\\
    \overline{Acc}, & \text{if }\overline{Acc}<0.5 \\
    WA, & \text{else}.
\end{cases}
\end{align}

\subsubsection{Score Modification}
In some cases it is beneficial to adapt the \ac{care}-score for different use cases. As the \ac{ad}-challenges \cite{hack_the_wind_2018} and \cite{we_do_wind_challenge_2021} show maintenance costs for different turbine faults are often considered when assessing the performance of \ac{ad}-models. The \ac{care}-score can be adjusted to take such costs into account by replacing $\overline{F_\beta}$ and $\overline{WS}$ by weighted averages. 

Let $a_1, \dots, a_N$ be all datasets containing anomaly events and $\boldsymbol{\omega}\coloneqq (\omega_1, \dots, \omega_N)$ be the cost-based importance weights of each anomaly. The weighted averages are then be defined by
\begin{align}
    \overline{F^{\boldsymbol{\omega}}_\beta}  &\coloneqq \frac{1}{\sum_{i=1}^N \omega_i}\sum_{i=1}^N\omega_i\cdot F_\beta(\mathbf{g}_{a_i}, \mathbf{p}_{a_i}) \\
    \overline{WS^{\boldsymbol{\omega}}} &\coloneqq \frac{1}{\sum_{i=1}^N \omega_i}\sum_{i=1}^N\omega_i\cdot WS(\mathbf{p}_{a_i}),
\end{align}
where $\mathbf{g}_{a_i}$ is the ground truth for the prediction time frame of dataset $a_i$ and $\mathbf{p}_{a_i}$ is the corresponding model prediction. Additionally, the weights in equation \ref{equation: weighted average} can be altered to better suit the use case.

For the following mini-benchmark section no further modification of weights are made, since the necessary information about maintenance costs for each fault in the dataset are not available for the wind farms B and C.

%% file: benchmark.tex
For the purpose of using the dataset for benchmarks of \ac{ad} algorithms for fault detection in \acp{wt}, and showcasing the newly defined \ac{care}-score, python implementations of an \ac{nbm} based on an \ac{ae} approach and a simple isolation forest approach are compared to the trivial strategies ``all anomaly", ``all normal" and ``random".

\subsubsection{Simple approaches}
While ``all anomaly" just classifies every timestamp as an anomaly and ``all normal" does the opposite, ``random" assigns the prediction for every timestamp independently based on a 50/50 probability. The slightly more complex isolation forest approach uses the implementation from the python package ``sklearn" \cite{scikit-learn} with ``n\_estimators"=100 and ``contamination"=0.09, as well as a principal component analysis in order to reduce the dimensionality of the input data, such that 99\% of the variance is kept. All hyperparameters were selected by manual tests.

\subsubsection{Autoencoder approach}
The \ac{ae} \ac{nbm} is a more sophisticated approach. The model used is a further developed version of the \ac{ad} procedure described in \cite{roelofs_autoencoder-based_2021}. This model consists of an \ac{ae} model trained on data representing normal behavior and a calibrated threshold to detect anomalies. The \ac{ae} models for each wind farm contain 3 to 5 hidden layers and are optimized using the Adam algorithm. The hyperparameters, such as the number of units in the hidden layers, the learning rate and the amount of noise to regularize the \ac{ae}, were adjusted using the python package ``Optuna" \cite{optuna_2019}. An overview of the model hyperparameters for the baseline models is provided in table \ref{table hyper parameters}. The \ac{ae} is trained on 75\% of the normal training data while 25\% of the data are randomly selected for validation. The training lasts at most 200 epochs with an early stopping option, which is triggered if the L2-norm of the reconstruction error on the validation data does not decrease for 3 consecutive epochs.
Based on a calibrated threshold, predicted timestamps are assigned the label ``anomaly" if the L2-norm of the corresponding reconstruction error exceeds the threshold, otherwise they will receive the label ``normal". 

The calibration of the threshold differs depending on the wind farm. For the wind farms A and B an adaptive threshold is used, inspired by the work in \cite{zhao_anomaly_2018} and \cite{chen_anomaly_2021}. Here, a \ac{nn} regression model is used to learn the mapping of the \ac{ae} input data to the L2-norm of \acp{re}. The \ac{nn} consists of 3 layers with around 20 to 40 units in the hidden layer, ReLU activations and the Adam algorithm is used for optimization. For training the same data which the \ac{ae} is validated on is used, i.e. the part of the validation data representing normal behavior. The training lasts at most 300 epochs with the same early stopping mechanism as in the \ac{ae} training. During prediction, the new input data is evaluated by the \ac{nn} and provides an expected \ac{re} $\epsilon$. This expected \ac{re} is increased by adding the sensitivity parameter $\gamma\in[0,\infty]$ and then compared to the actual \ac{re} of the \ac{ae} model. While parameter $\gamma$ has to be optimized for each wind farm separately, values from the interval $[0.2,0.4]$ seem to be a good fit for the provided datasets. If the actual \ac{re} is larger than $\epsilon + \gamma$, the corresponding timestamp is detected as an ``anomaly''. For the determination of the optimal number of units in the hidden layer of the \ac{nn} and the value for $\gamma$ a hyperparameter optimization with ``Optuna" is used. The final hyperparameter values of the thresholds used for the benchmark can be found in table \ref{table hyper parameters}.

For wind farm C, a fixed threshold is calibrated. For this, the L2-norm of the reconstruction errors (anomaly score) of all \ac{ae} validation data is computed. Afterwards, the constant threshold is selected by iterating over the calculated anomaly score values and choosing the value that maximizes the $F_\frac{1}{2}$-score based the ground truth defined by the normal behavior labels, that can be derived from the status-IDs as shown in table \ref{table status}.

\begin{center}
\begin{tabularx}{\textwidth}{
    >{\centering\arraybackslash}X
  | >{\centering\arraybackslash}X 
  | >{\centering\arraybackslash}X
  | >{\centering\arraybackslash}X
} \label{table hyper parameters}
    & Wind Farm A & Wind Farm B & Wind Farm C  \\
    \hline
    \# units in hidden layers & 44, 25, 4, 25, 44 & 40, 15, 40 & 133, 83, 20, 83, 133 \\
    \hline
    learning rate & 0.0018 & 0.003 & 0.0056 \\
    \hline
    noise & 0.06 & 0 & 0 \\
    \hline
    batch size & 64 & 64 & 64 \\
    \hline
    threshold & adaptive & adaptive & max. $F_\frac{1}{2}$-score \\
    \hline
    threshold \ac{nn} hidden layers & 23 & 36 & - \\
    \hline
    $\gamma$ & 0.344 & 0.234 & -
\end{tabularx}
\end{center}

Finally the \ac{ae} \ac{nbm} is supplemented with an additional data filter. In order to remove potentially implausible data from the training data of the \ac{ae} a status based on wind speed and power enhances the normal behavior labels given by the turbine operational status from table \ref{table status}. For the determination of the new status information timestamps are marked as not normal if the wind speed is within the normal operation range of the \ac{wt} and the power is close or equal to 0.

\subsubsection{Scoring of the approaches}
\begin{figure}
    \centering
    \includegraphics[width=\textwidth]{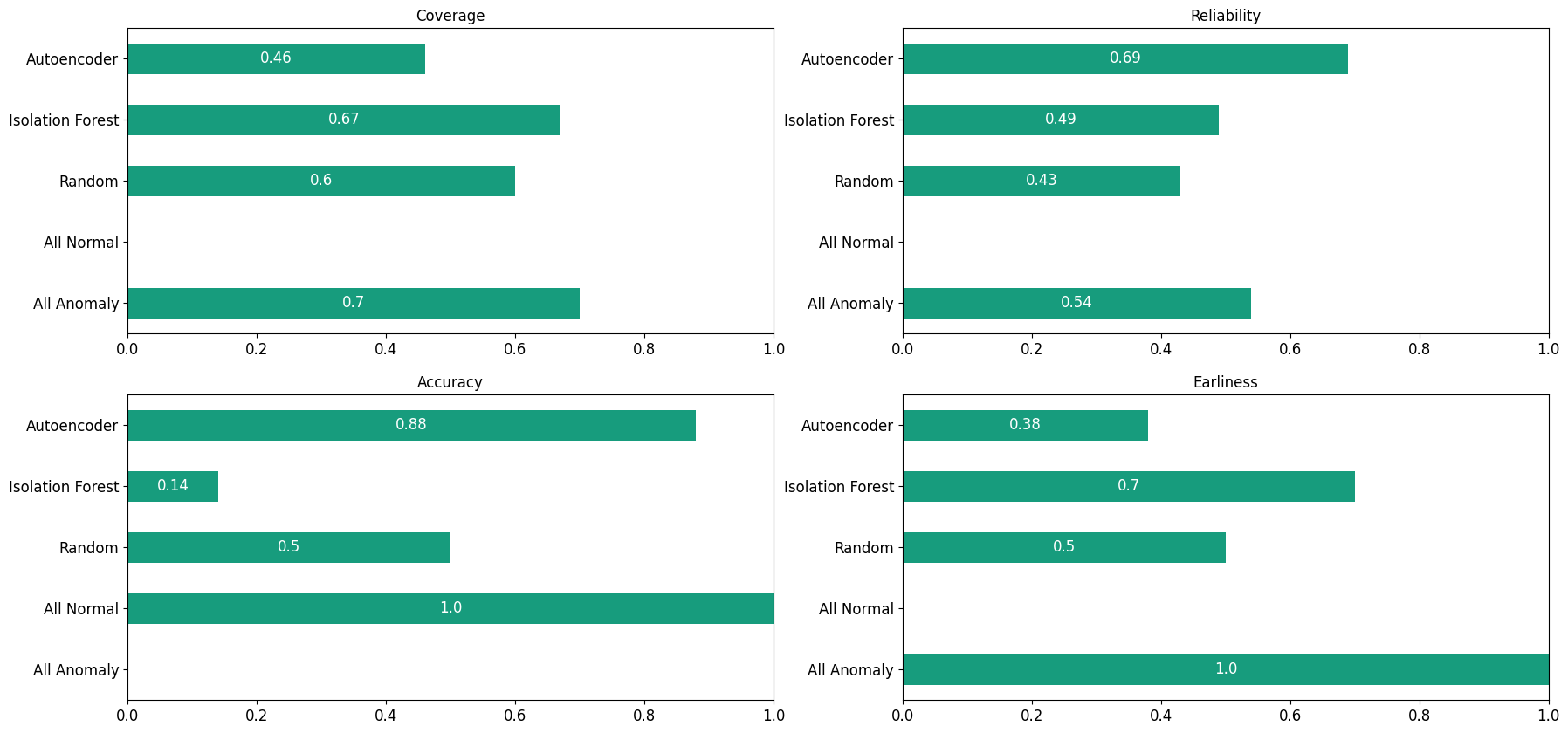}
    \caption{Sub-scores of \ac{care}-score over all 95 sub-datasets for a few selected approaches.}
    \label{figure sub scores}
\end{figure}

At first the four sub-scores of the \ac{care}-score are evaluated for all five approaches. The results are visualized in figure \ref{figure sub scores}. While ``all anomaly" obviously performs well in detecting anomalies, it of course has the worst possible accuracy on normal data. The opposite is of course the case for ``all normal". The isolation forest behaves similar to ``all anomaly", since it detects a lot of anomalies but performed very poorly in recognizing normal behavior. Finally the \ac{ae} approach has a high accuracy on normal data and a good performance on the event based $F_\frac{1}{2}$-score but it suffers a little bit from an overall low number of detected anomalies.

\begin{figure}
    \centering
    \includegraphics[width=\textwidth]{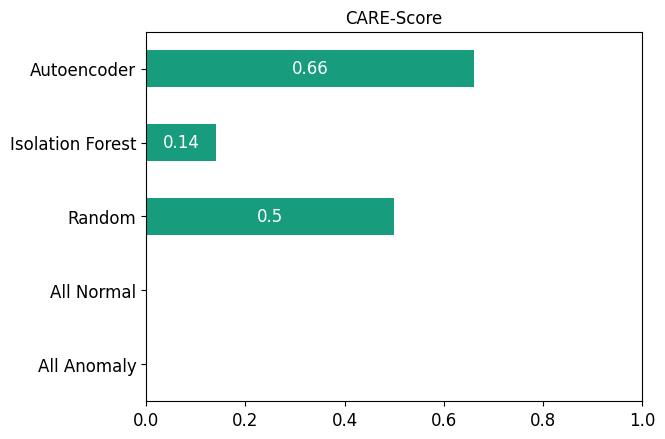}
    \caption{\ac{care}-score benchmark over all 95 sub-datasets for a few selected approaches.}
    \label{figure final score}
\end{figure}

When it comes to the final \ac{care}-score shown in figure \ref{figure final score}, the trivial strategies ``all anomaly" and ``all normal" both get the score 0 because they run into the special cases described at the end of section \ref{section score}. The strategy ``random" gets the \ac{care}-score of 0.5 and sets the lower bound to beat for good any anomaly detection algorithms. The isolation forest approach does not outperform that threshold since it is not able to recognize normal behavior appropriately with the chosen parameter configuration. With a score of 0.66 the \ac{ae} approach represents a good anomaly detector since it is able to detect anomalies while also recognizing normal behavior very well and having a good classification performance on aggregated events.

%% file: summary.tex
With the purpose of reducing the limitations that come with the lack of public data for \ac{wt} \ac{ad} benchmarks, a new dataset was published. Composed out of multiple \acp{wt} across 3 wind farms the dataset shows greater detail in anomaly labels and additional information than datasets that are currently available. By formulating requirements for benchmark datasets for \ac{ad} in \acp{wt}, the data quality and the ability to test for generalization of \ac{ad} models were ensured. 
The balanced nature of the dataset, with similar amounts of anomalous data and examples of normal behavior, allows for more detailed and meaningful comparison studies of \ac{ad} algorithms.

Furthermore, we proposed an evaluation method, the \ac{care}-score, that fully uses the informational depth the dataset provided. By considering the four key aspects of a good \ac{ad}-model 
- detecting many anomalies, early detection, few false alarms and correct recognition of normal behavior - the \ac{care}-score provides a measure for the all-around performance of a good \ac{ad} model for predictive maintenance on \acp{wt}.

To demonstrate the combination of dataset and \ac{care}-score, a `mini-benchmark' was conducted. A sophisticated \ac{ad} algorithm was compared to the popular and simpler isolation forest and 3 trivial strategies. This evaluation shows the importance of not neglecting normal behavior recognition while trying to  detect as many anomalies as possible, as it was the case for the isolation forest approach.

As subject to future research the provided dataset and scoring method can be used to compare a wide range of \ac{wt} \ac{ad} models on an equal and transparent basis in order to push the progress in this field further and find good \ac{ad} algorithms.

To further enhance the development of benchmark datasets in the field of \ac{wt} \ac{ad}, the authors strongly encourage others to share their data. Increasing the availability of high-quality benchmark datasets will facilitate more comprehensive and rigorous evaluations of \ac{ad} models in this domain.

%% file: ms.bbl
\begin{thebibliography}{10}
\expandafter\ifx\csname url\endcsname\relax
  \def\url#1{\texttt{#1}}\fi
\expandafter\ifx\csname urlprefix\endcsname\relax\def\urlprefix{URL }\fi
\expandafter\ifx\csname href\endcsname\relax
  \def\href#1#2{#2} \def\path#1{#1}\fi

\bibitem{tautzweinert_using_2017}
J.~Tautz‐Weinert, S.~J. Watson,
  \href{https://onlinelibrary.wiley.com/doi/10.1049/iet-rpg.2016.0248}{Using
  {SCADA} data for wind turbine condition monitoring – a review}, IET
  Renewable Power Generation 11~(4) (2017) 382--394.
\newblock \href {https://doi.org/10.1049/iet-rpg.2016.0248}
  {\path{doi:10.1049/iet-rpg.2016.0248}}.
\newline\urlprefix\url{https://onlinelibrary.wiley.com/doi/10.1049/iet-rpg.2016.0248}

\bibitem{helbing_deep_2018}
G.~Helbing, M.~Ritter,
  \href{https://www.sciencedirect.com/science/article/pii/S1364032118306610}{Deep
  {Learning} for fault detection in wind turbines}, Renewable and Sustainable
  Energy Reviews 98 (2018) 189--198.
\newblock \href {https://doi.org/10.1016/j.rser.2018.09.012}
  {\path{doi:10.1016/j.rser.2018.09.012}}.
\newline\urlprefix\url{https://www.sciencedirect.com/science/article/pii/S1364032118306610}

\bibitem{pandit2024}
R.~Pandit, J.~Wang,
  \href{https://ietresearch.onlinelibrary.wiley.com/doi/abs/10.1049/rpg2.12920}{A
  comprehensive review on enhancing wind turbine applications with advanced
  {SCADA} data analytics and practical insights}, IET Renewable Power
  Generation 18~(4) (2024) 722--742, \_eprint:
  https://ietresearch.onlinelibrary.wiley.com/doi/pdf/10.1049/rpg2.12920.
\newblock \href {https://doi.org/https://doi.org/10.1049/rpg2.12920}
  {\path{doi:https://doi.org/10.1049/rpg2.12920}}.
\newline\urlprefix\url{https://ietresearch.onlinelibrary.wiley.com/doi/abs/10.1049/rpg2.12920}

\bibitem{latiffianti_wind_2022}
E.~Latiffianti, S.~Sheng, Y.~Ding,
  \href{https://www.frontiersin.org/articles/10.3389/fen rg.2022.904622}{Wind
  {Turbine} {Gearbox} {Failure} {Detection} {Through} {Cumulative} {Sum} of
  {Multivariate} {Time} {Series} {Data}}, Frontiers in Energy Research 10
  (2022).
\newblock \href {https://doi.org/10.3389/fenrg.2022.904622}
  {\path{doi:10.3389/fenrg.2022.904622}}.
\newline\urlprefix\url{https://www.frontiersin.org/articles/10.3389/fen
  rg.2022.904622}

\bibitem{adbench2022}
S.~Han, X.~Hu, H.~Huang, M.~Jiang, Y.~Zhao,
  \href{https://proceedings.neurips.cc/paper_files/paper/2022/file/cf93972b116ca5268827d575f2cc226b-Paper-Datasets_and_Benchmarks.pdf}{{ADBench}:
  {Anomaly} {Detection} {Benchmark}}, in: S.~Koyejo, S.~Mohamed, A.~Agarwal,
  D.~Belgrave, K.~Cho, A.~Oh (Eds.), Advances in {Neural} {Information}
  {Processing} {Systems}, Vol.~35, Curran Associates, Inc., 2022, pp.
  32142--32159.
\newline\urlprefix\url{https://proceedings.neurips.cc/paper_files/paper/2022/file/cf93972b116ca5268827d575f2cc226b-Paper-Datasets_and_Benchmarks.pdf}

\bibitem{numenta2015}
A.~Lavin, S.~Ahmad, Evaluating real-time anomaly detection algorithms -- the
  numenta anomaly benchmark, in: 2015 IEEE 14th International Conference on
  Machine Learning and Applications (ICMLA), 2015, pp. 38--44.
\newblock \href {https://doi.org/10.1109/ICMLA.2015.141}
  {\path{doi:10.1109/ICMLA.2015.141}}.

\bibitem{pang2022}
G.~Pang, C.~Shen, L.~Cao, A.~V.~D. Hengel,
  \href{https://dl.acm.org/doi/10.1145/3439950}{Deep {Learning} for {Anomaly}
  {Detection}: {A} {Review}}, ACM Computing Surveys 54~(2) (2022) 1--38.
\newblock \href {https://doi.org/10.1145/3439950} {\path{doi:10.1145/3439950}}.
\newline\urlprefix\url{https://dl.acm.org/doi/10.1145/3439950}

\bibitem{schmidl_anomaly_2022}
S.~Schmidl, P.~Wenig, T.~Papenbrock,
  \href{https://doi.org/10.14778/3538598.3538602}{Anomaly detection in time
  series: a comprehensive evaluation}, Proc. VLDB Endow. 15~(9) (2022)
  1779--1797, publisher: VLDB Endowment.
\newblock \href {https://doi.org/10.14778/3538598.3538602}
  {\path{doi:10.14778/3538598.3538602}}.
\newline\urlprefix\url{https://doi.org/10.14778/3538598.3538602}

\bibitem{zhang_research_2024}
C.~Zhang, D.~Hu, T.~Yang,
  \href{https://www.sciencedirect.com/science/article/pii/S0951832023005483}{Research
  of artificial intelligence operations for wind turbines considering anomaly
  detection, root cause analysis, and incremental training}, Reliability
  Engineering \& System Safety 241 (2024) 109634.
\newblock \href {https://doi.org/https://doi.org/10.1016/j.ress.2023.109634}
  {\path{doi:https://doi.org/10.1016/j.ress.2023.109634}}.
\newline\urlprefix\url{https://www.sciencedirect.com/science/article/pii/S0951832023005483}

\bibitem{yang_luoxiao_conditional_2021}
{Yang Luoxiao}, {Zhang Zijun}, A {Conditional} {Convolutional}
  {Autoencoder}-{Based} {Method} for {Monitoring} {Wind} {Turbine} {Blade}
  {Breakages}, IEEE Transactions on Industrial Informatics 17~(9) (2021)
  6390--6398.
\newblock \href {https://doi.org/10.1109/TII.2020.3011441}
  {\path{doi:10.1109/TII.2020.3011441}}.

\bibitem{morrison_anomaly_2022}
R.~Morrison, X.~Liu, Z.~Lin,
  \href{https://www.sciencedirect.com/science/article/pii/S0960148121017134}{Anomaly
  detection in wind turbine {SCADA} data for power curve cleaning}, Renewable
  Energy 184 (2022) 473--486.
\newblock \href {https://doi.org/https://doi.org/10.1016/j.renene.2021.11.118}
  {\path{doi:https://doi.org/10.1016/j.renene.2021.11.118}}.
\newline\urlprefix\url{https://www.sciencedirect.com/science/article/pii/S0960148121017134}

\bibitem{schroder_using_2022}
L.~Schröder, N.~K. Dimitrov, D.~R. Verelst, J.~A. Sørensen,
  \href{https://www.mdpi.com/1996-1073/15/2/558}{Using {Transfer} {Learning} to
  {Build} {Physics}-{Informed} {Machine} {Learning} {Models} for {Improved}
  {Wind} {Farm} {Monitoring}}, Energies 15~(2) (2022).
\newblock \href {https://doi.org/10.3390/en15020558}
  {\path{doi:10.3390/en15020558}}.
\newline\urlprefix\url{https://www.mdpi.com/1996-1073/15/2/558}

\bibitem{mckinnon_comparison_2020}
C.~McKinnon, J.~Carroll, A.~McDonald, S.~Koukoura, D.~Infield, C.~Soraghan,
  \href{https://www.mdpi.com/1996-1073/13/19/5152}{Comparison of {New}
  {Anomaly} {Detection} {Technique} for {Wind} {Turbine} {Condition}
  {Monitoring} {Using} {Gearbox} {SCADA} {Data}}, Energies 13~(19) (2020).
\newblock \href {https://doi.org/10.3390/en13195152}
  {\path{doi:10.3390/en13195152}}.
\newline\urlprefix\url{https://www.mdpi.com/1996-1073/13/19/5152}

\bibitem{jia_condition_2021}
X.~Jia, Y.~Han, Y.~Li, Y.~Sang, G.~Zhang, Condition monitoring and performance
  forecasting of wind turbines based on denoising autoencoder and novel
  convolutional neural networks, Energy Reports 7 (2021) 6354--6365.
\newblock \href {https://doi.org/10.1016/j.egyr.2021.09.080}
  {\path{doi:10.1016/j.egyr.2021.09.080}}.

\bibitem{de_sa_wind_2020}
F.~P.~G. de~Sá, D.~N. Brandão, E.~Ogasawara, R.~d.~C. Coutinho, R.~F. Toso,
  Wind {Turbine} {Fault} {Detection}: {A} {Semi}-{Supervised} {Learning}
  {Approach} {With} {Automatic} {Evolutionary} {Feature} {Selection}, in: 2020
  {International} {Conference} on {Systems}, {Signals} and {Image} {Processing}
  ({IWSSIP}), 2020, pp. 323--328.
\newblock \href {https://doi.org/10.1109/IWSSIP48289.2020.9145244}
  {\path{doi:10.1109/IWSSIP48289.2020.9145244}}.

\bibitem{udo_data-driven_2021}
W.~Udo, Y.~Muhammad, Data-{Driven} {Predictive} {Maintenance} of {Wind}
  {Turbine} {Based} on {SCADA} {Data}, IEEE Access 9 (2021) 162370--162388.
\newblock \href {https://doi.org/10.1109/ACCESS.2021.3132684}
  {\path{doi:10.1109/ACCESS.2021.3132684}}.

\bibitem{tang_fault_2023}
Z.~Tang, X.~Shi, H.~Zou, Y.~Zhu, Y.~Yang, Y.~Zhang, J.~He,
  \href{https://www.mdpi.com/1424-8220/23/13/6198}{Fault {Diagnosis} of {Wind}
  {Turbine} {Generators} {Based} on {Stacking} {Integration} {Algorithm} and
  {Adaptive} {Threshold}}, Sensors 23~(13) (2023).
\newblock \href {https://doi.org/10.3390/s23136198}
  {\path{doi:10.3390/s23136198}}.
\newline\urlprefix\url{https://www.mdpi.com/1424-8220/23/13/6198}

\bibitem{jankauskas_exploring_2023}
M.~Jankauskas, A.~Serackis, M.~Šapurov, R.~Pomarnacki, A.~Baskys, V.~K. Hyunh,
  T.~Vaimann, J.~Zakis,
  \href{https://www.mdpi.com/1424-8220/23/12/5695}{Exploring the {Limits} of
  {Early} {Predictive} {Maintenance} in {Wind} {Turbines} {Applying} an
  {Anomaly} {Detection} {Technique}}, Sensors 23~(12) (2023).
\newblock \href {https://doi.org/10.3390/s23125695}
  {\path{doi:10.3390/s23125695}}.
\newline\urlprefix\url{https://www.mdpi.com/1424-8220/23/12/5695}

\bibitem{barber_best_2023}
S.~Barber, U.~Izagirre, O.~Serradilla, J.~Olaizola, E.~Zugasti, J.~I. Aizpurua,
  A.~E. Milani, F.~Sehnke, Y.~Sakagami, C.~Henderson,
  \href{https://www.mdpi.com/1996-1073/16/8/3567}{Best {Practice} {Data}
  {Sharing} {Guidelines} for {Wind} {Turbine} {Fault} {Detection} {Model}
  {Evaluation}}, Energies 16~(8) (2023).
\newblock \href {https://doi.org/10.3390/en16083567}
  {\path{doi:10.3390/en16083567}}.
\newline\urlprefix\url{https://www.mdpi.com/1996-1073/16/8/3567}

\bibitem{barber_enabling_2022}
S.~Barber, L.~A.~M. Lima, Y.~Sakagami, J.~Quick, E.~Latiffianti, Y.~Liu,
  R.~Ferrari, S.~Letzgus, X.~Zhang, F.~Hammer,
  \href{https://www.mdpi.com/1996-1073/15/15/5638}{Enabling {Co}-{Innovation}
  for a {Successful} {Digital} {Transformation} in {Wind} {Energy} {Using} a
  {New} {Digital} {Ecosystem} and a {Fault} {Detection} {Case} {Study}},
  Energies 15~(15) (2022).
\newblock \href {https://doi.org/10.3390/en15155638}
  {\path{doi:10.3390/en15155638}}.
\newline\urlprefix\url{https://www.mdpi.com/1996-1073/15/15/5638}

\bibitem{nassif_machine_2021}
A.~B. Nassif, M.~A. Talib, Q.~Nasir, F.~M. Dakalbab, Machine {Learning} for
  {Anomaly} {Detection}: {A} {Systematic} {Review}, IEEE Access 9 (2021)
  78658--78700.
\newblock \href {https://doi.org/10.1109/ACCESS.2021.3083060}
  {\path{doi:10.1109/ACCESS.2021.3083060}}.

\bibitem{ruff_unifying_2021}
L.~Ruff, J.~R. Kauffmann, R.~A. Vandermeulen, G.~Montavon, W.~Samek, M.~Kloft,
  T.~G. Dietterich, K.-R. Muller,
  \href{https://ieeexplore.ieee.org/document/9347460/}{A {Unifying} {Review} of
  {Deep} and {Shallow} {Anomaly} {Detection}}, Proceedings of the IEEE 109~(5)
  (2021) 756--795.
\newblock \href {https://doi.org/10.1109/JPROC.2021.3052449}
  {\path{doi:10.1109/JPROC.2021.3052449}}.
\newline\urlprefix\url{https://ieeexplore.ieee.org/document/9347460/}

\bibitem{effenberger_collection_2022}
N.~Effenberger, N.~Ludwig,
  \href{https://onlinelibrary.wiley.com/doi/abs/10.1002/we.2766}{A collection
  and categorization of open-source wind and wind power datasets}, Wind Energy
  25~(10) (2022) 1659--1683, \_eprint:
  https://onlinelibrary.wiley.com/doi/pdf/10.1002/we.2766.
\newblock \href {https://doi.org/https://doi.org/10.1002/we.2766}
  {\path{doi:https://doi.org/10.1002/we.2766}}.
\newline\urlprefix\url{https://onlinelibrary.wiley.com/doi/abs/10.1002/we.2766}

\bibitem{menezes_wind_2020}
D.~Menezes, M.~Mendes, J.~A. Almeida, T.~Farinha,
  \href{https://www.mdpi.com/1996-1073/13/18/4702}{Wind {Farm} and {Resource}
  {Datasets}: {A} {Comprehensive} {Survey} and {Overview}}, Energies 13~(18)
  (2020).
\newblock \href {https://doi.org/10.3390/en13184702}
  {\path{doi:10.3390/en13184702}}.
\newline\urlprefix\url{https://www.mdpi.com/1996-1073/13/18/4702}

\bibitem{letzgus_wind_nodate}
S.~Letzgus,
  \href{https://github.com/sltzgs/Wind_Turbine_SCADA_open_data?tab=readme-ov-file}{Wind
  {Turbine} {SCADA} open data}, [Online; accessed 13-03-2024] (2023).
\newline\urlprefix\url{https://github.com/sltzgs/Wind_Turbine_SCADA_open_data?tab=readme-ov-file}

\bibitem{edp_inovacao_edpr_2018}
{EDP Inovação},
  \href{https://www.edp.com/en/innovation/open-data/data}{\textsc{{EDPR}}
  {Wind} {Farm} {Open} {Data}: {Wind} {Turbine} {SCADA} signals and historical
  failure logbook from 2016 and 2017} (2018).
\newline\urlprefix\url{https://www.edp.com/en/innovation/open-data/data}

\bibitem{hack_the_wind_2018}
{EDP Inovação},
  \href{https://www.edp.com/en/innovation/open-data/reuses/hack-the-wind}{{Hack}
  {the} {Wind}: {Wind} {Turbine} {Failures} {Detection}} (2018).
\newline\urlprefix\url{https://www.edp.com/en/innovation/open-data/reuses/hack-the-wind}

\bibitem{we_do_wind_challenge_2021}
{Eastern Switzerland University of Applied Sciences},
  \href{https://www.wedowind.ch/spaces/edp-challenges-space}{{Wo} {do} {Wind}:
  {EDP} {Challenges} {space}} (2021).
\newline\urlprefix\url{https://www.wedowind.ch/spaces/edp-challenges-space}

\bibitem{renjie2023}
R.~Wu, E.~J. Keogh, Current time series anomaly detection benchmarks are flawed
  and are creating the illusion of progress, IEEE Transactions on Knowledge and
  Data Engineering 35~(3) (2023) 2421--2429.
\newblock \href {https://doi.org/10.1109/TKDE.2021.3112126}
  {\path{doi:10.1109/TKDE.2021.3112126}}.

\bibitem{chen_anomaly_2021}
H.~Chen, H.~Liu, X.~Chu, Q.~Liu, D.~Xue,
  \href{https://www.sciencedirect.com/science/article/pii/S0960148121004341}{Anomaly
  detection and critical {SCADA} parameters identification for wind turbines
  based on {LSTM}-{AE} neural network}, Renewable Energy 172 (2021) 829--840.
\newblock \href {https://doi.org/https://doi.org/10.1016/j.renene.2021.03.078}
  {\path{doi:https://doi.org/10.1016/j.renene.2021.03.078}}.
\newline\urlprefix\url{https://www.sciencedirect.com/science/article/pii/S0960148121004341}

\bibitem{astha2022}
A.~Garg, W.~Zhang, J.~Samaran, R.~Savitha, C.-S. Foo, An evaluation of anomaly
  detection and diagnosis in multivariate time series, IEEE Transactions on
  Neural Networks and Learning Systems 33~(6) (2022) 2508--2517.
\newblock \href {https://doi.org/10.1109/TNNLS.2021.3105827}
  {\path{doi:10.1109/TNNLS.2021.3105827}}.

\bibitem{carrasco2021}
J.~Carrasco, D.~López, I.~Aguilera-Martos, D.~García-Gil, I.~Markova,
  M.~García-Barzana, M.~Arias-Rodil, J.~Luengo, F.~Herrera,
  \href{https://www.sciencedirect.com/science/article/pii/S0925231221011826}{Anomaly
  detection in predictive maintenance: A new evaluation framework for temporal
  unsupervised anomaly detection algorithms}, Neurocomputing 462 (2021)
  440--452.
\newblock \href {https://doi.org/https://doi.org/10.1016/j.neucom.2021.07.095}
  {\path{doi:https://doi.org/10.1016/j.neucom.2021.07.095}}.
\newline\urlprefix\url{https://www.sciencedirect.com/science/article/pii/S0925231221011826}

\bibitem{stetco_machine_2019}
A.~Stetco, F.~Dinmohammadi, X.~Zhao, V.~Robu, D.~Flynn, M.~Barnes, J.~Keane,
  G.~Nenadic,
  \href{https://linkinghub.elsevier.com/retrieve/pii/S096014811831231X}{Machine
  learning methods for wind turbine condition monitoring: {A} review},
  Renewable Energy 133 (2019) 620--635.
\newblock \href {https://doi.org/10.1016/j.renene.2018.10.047}
  {\path{doi:10.1016/j.renene.2018.10.047}}.
\newline\urlprefix\url{https://linkinghub.elsevier.com/retrieve/pii/S096014811831231X}

\bibitem{scikit-learn}
F.~Pedregosa, G.~Varoquaux, A.~Gramfort, V.~Michel, B.~Thirion, O.~Grisel,
  M.~Blondel, P.~Prettenhofer, R.~Weiss, V.~Dubourg, J.~Vanderplas, A.~Passos,
  D.~Cournapeau, M.~Brucher, M.~Perrot, E.~Duchesnay, Scikit-learn: Machine
  learning in {P}ython, Journal of Machine Learning Research 12 (2011)
  2825--2830.

\bibitem{roelofs_autoencoder-based_2021}
C.~M. Roelofs, M.-A. Lutz, S.~Faulstich, S.~Vogt, Autoencoder-based anomaly
  root cause analysis for wind turbines, Energy and AI 4 (2021) 100065.
\newblock \href {https://doi.org/10.1016/j.egyai.2021.100065}
  {\path{doi:10.1016/j.egyai.2021.100065}}.

\bibitem{optuna_2019}
T.~Akiba, S.~Sano, T.~Yanase, T.~Ohta, M.~Koyama, Optuna: A next-generation
  hyperparameter optimization framework, in: Proceedings of the 25th {ACM}
  {SIGKDD} International Conference on Knowledge Discovery and Data Mining,
  2019, p. 2623–2631.

\bibitem{zhao_anomaly_2018}
H.~Zhao, H.~Liu, W.~Hu, X.~Yan,
  \href{https://www.sciencedirect.com/science/article/pii/S0960148118305457}{Anomaly
  detection and fault analysis of wind turbine components based on deep
  learning network}, Renewable Energy 127 (2018) 825--834.
\newblock \href {https://doi.org/10.1016/j.renene.2018.05.024}
  {\path{doi:10.1016/j.renene.2018.05.024}}.
\newline\urlprefix\url{https://www.sciencedirect.com/science/article/pii/S0960148118305457}

\end{thebibliography}
